\documentclass[letterpaper]{article} %
\usepackage[submission]{aaai24}  %
\usepackage{times}  %
\usepackage{helvet}  %
\usepackage{courier}  %
\usepackage[hyphens]{url}  %
\usepackage{graphicx} %
\def\UrlFont{\rm}  %
\usepackage{natbib}  %
\usepackage{caption} %
\usepackage{algorithm}
\usepackage{algorithmic}
\usepackage[utf8]{inputenc} %
\usepackage[T1]{fontenc}    %
\usepackage{url}            %
\usepackage{booktabs}       %
\usepackage{amsfonts}       %
\usepackage{nicefrac}       %
\usepackage{microtype}      %
\usepackage{xcolor}         %
\usepackage{graphicx}
\usepackage{subfigure}
\usepackage{multirow}
\usepackage{makecell}
\usepackage{enumitem}
\usepackage{bm}
\usepackage{hyperref}        %

\usepackage{newfloat}
\usepackage{listings}
\DeclareCaptionStyle{ruled}{labelfont=normalfont,labelsep=colon,strut=off} %
\floatstyle{ruled}
\newfloat{listing}{tb}{lst}{}
\floatname{listing}{Listing}
\title{Improving the Stability of GNN Force Field Models \\ by Reducing Feature Correlation}
\author {
    Yujie Zeng\textsuperscript{\rm 1},
    Wenlong He\textsuperscript{\rm 1}, 
    Ihor Vasyltsov\textsuperscript{\rm 2},\\
    Jiaxin Wei\textsuperscript{\rm 1},
    Ying Zhang\textsuperscript{\rm 1},
    Lin Chen\textsuperscript{\rm 1},
    Yuehua Dai\textsuperscript{\rm 1}
}
\affiliations {
    \textsuperscript{\rm 1} Samsung Research Institute China Xian, Xian, 710000, China \\
    \textsuperscript{\rm 2} Samsung Advanced Institute of Technology Suwon-si, Gyeonggi-do, 16678, Korea \\
    \texttt{\{yujie.zeng, wenlong.he, ihor.vasiltsov\}@samsumg.com}\\
    \texttt{\{jiaixn.wei, ying07.zhang, lin81.chen, yuehua.dai\}@samsumg.com}\\
}

\begin{document}

\maketitle

\begin{abstract}
    Recently, Graph Neural Network based Force Field (GNNFF) models are widely used in Molecular Dynamics (MD) simulation, which is one of the most cost-effective means in semiconductor material research. However, even such models provide high accuracy in energy and force Mean Absolute Error (MAE) over trained (in-distribution) datasets, they often become unstable during long-time MD simulation when used for out-of-distribution datasets. In this paper, we propose a feature correlation based method for GNNFF models to enhance the stability of MD simulation. We reveal the negative relationship between feature correlation and the stability of GNNFF models, and design a loss function with a dynamic loss coefficient scheduler to reduce edge feature correlation that can be applied in general GNNFF training. We also propose an empirical metric to evaluate the stability in MD simulation. Experiments show our method can significantly improve stability for GNNFF models especially in out-of-distribution data with less than 3\% computational overhead. For example, we can ensure the stable MD simulation time from 0.03ps to 10ps for Allegro model.
\end{abstract}

\section{Introduction}

The development and innovation of semiconductor devices rely deeply on the study of semiconductor material properties \citep{kim_review_2022, redaelli_historical_2022, orji_metrology_2019, nakamae_electron_2021}. This research requires accurate and effective experiments and visualization of atomic scale interaction and formation. However, natural experiments are costly and time-consuming. Thus, Molecular Dynamics simulation has emerged to be a cost-effective way to study material properties and reduce detrimental defects in semiconductor materials area \citep{gu_molecular_2022, zhou_impact_2019}.
MD simulation is a widely used theoretical method to simulate the motion of a system of interacting particles such as atoms. It can represent the simulation results of nanomaterials depending on the availability of proper potential functions/force fields modeling interatomic forces \citep{thompson_lammps_2022, alder_studies_1959, rahman_correlations_1964, frenkel_chapter_2002}. These results are useful in laboratory and industrial applications in material and biology science. 

Various Force Fields (FF) models were developed to study different aspects of material properties \citep{gu_molecular_2022, zhou_impact_2019}. Classical FF can be obtained from first principles using a quantum mechanical method such as Density Functional Theory (DFT) \citep{van_mourik_density_2014}. This is called Ab Inito MD (AIMD). AIMD can provide extremely high accuracy with theoretical considerations rather than empirical fitting \citep{iftimie_ab_2005}. However, the significant disadvantage of AIMD is that it calculates the potential with treating the electronic degrees of freedom, therefore it's limited to short simulations due to the huge computation cost. Moreover, AIMD is limited to systems that contain several hundreds of atoms. However, the demand for large-scale atom system simulations in industry has been increasing recently. Accordingly, more and more Machine Learning (ML) and Deep Learning (DL) methods are researched and applied in MD area \citep{anstine_machine_2023, jia_pushing_2020} due to the high accuracy and better scalability for large atomic systems. Among these ML based Force Field (MLFF) models, Graph Neural Networks based Force Field (GNNFF) models have shown its ability to capture the atomic interaction with graph-based system modeling \citep{batzner_e3-equivariant_2022, musaelian_learning_2023, gasteiger_gemnet_2021, schutt_schnet_2017, mailoa_fast_2019, park_accurate_2021}. 
GNNFF models take particle position, particle features and spatial features as input to model the interactions of atoms and learn to predict particle energy and forces of the whole system. The predicted forces then are used in MD simulation tools (e.g., LAMMPS \citep{thompson_lammps_2022}) to calculate particle positions after a time step. Recently, many GNNFF models are developed and used, such as NequIP \citep{batzner_e3-equivariant_2022}, Allegro \citep{musaelian_learning_2023}, GemNet \citep{gasteiger_gemnet_2021} and SCHNet \citep{schutt_schnet_2017}. In this paper, we mainly focus on NequIP, Allegro and GemNet models because of their superior accuracy and scalability.

\begin{figure*}[t]
  \centering
  \begin{minipage}[t]{0.48\textwidth}
    \centering
    \includegraphics[width=\textwidth,height=0.48\textwidth]{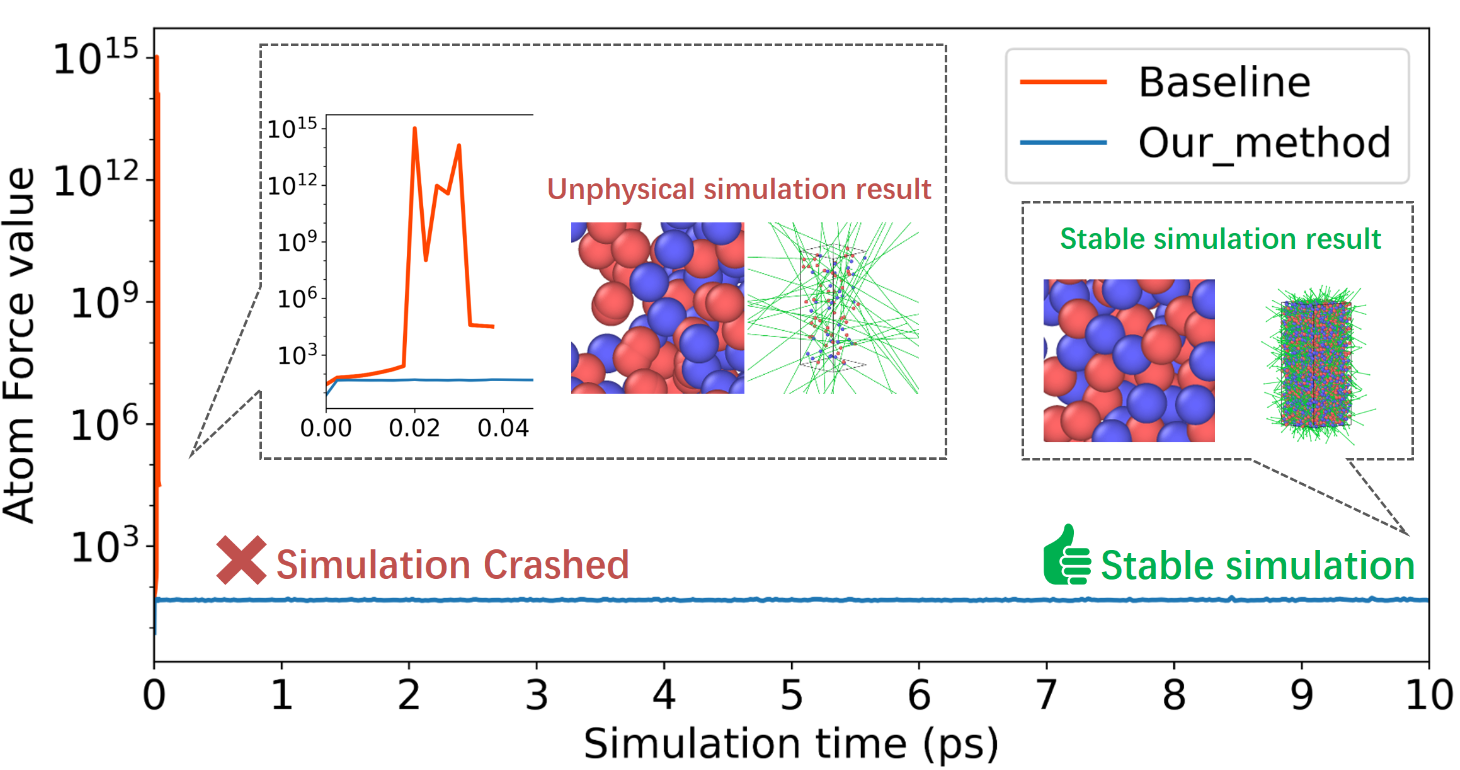}
    \vspace*{-1.0\baselineskip}

    \caption{MD simulation result with baseline Allegro model and our method}
    \label{fig_nonphysical}
  \end{minipage} \hfill
  \begin{minipage}[t]{0.48\textwidth}
    \centering
    \includegraphics[width=\textwidth,height=0.48\textwidth]{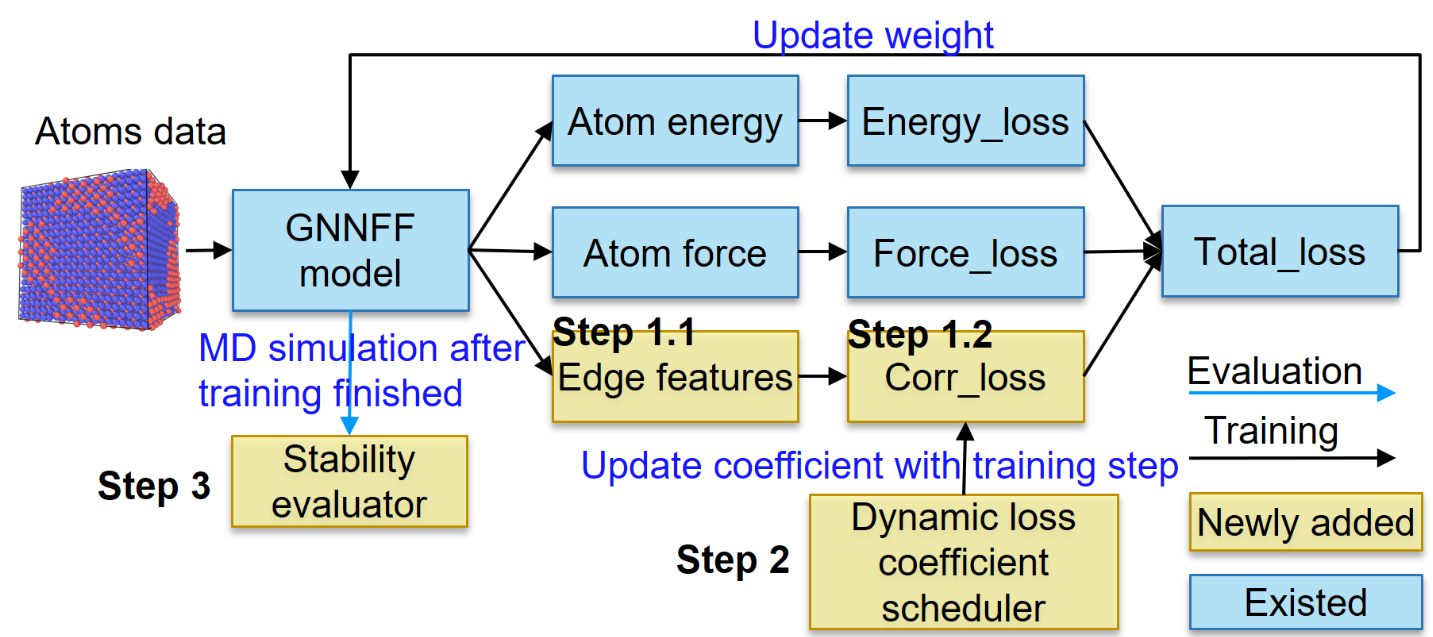}
    \vspace*{-1.0\baselineskip}

    \caption{General flow of the proposed method}
    \label{fig_overview}
  \end{minipage}

  \vspace*{-1.0\baselineskip}
\end{figure*}

\section{Motivation and key contributions}

Generally, the accuracy of atom energy and forces is the most important metric for GNNFF model since a more accurate prediction result can better reveal the macroscopic properties for materials and provide valuable insights to users. However, a model with good accuracy value in energy/force MAEs\footnote{Energy MAE and Force MAE are typical metric used to estimate the accuracy of the FF models. \citep{geonu_kim_benchmark_2023}} cannot ensure stability since the accuracy value only guarantees that trained model learns the knowledge from the training data, which often can be incomplete, or biased. Thus, simulation stability is an important challenge to MD simulation methods especially in long-time simulations \citep{stocker_how_2022, fu_forces_2022, bihani_egraffbench_2023, fu_simulate_2023}. GNNFF models may produce unstable or wrong prediction result when the learned force field is not robust enough. The simulation can enter nonphysical states and MD simulation will end up as system crash as shown in Figure \ref{fig_nonphysical}. Therefore, improving the stability of GNNFF model is important in real application scenarios.

Besides, in real application scenarios of MD simulation, GNNFF models are expected to be robust in as many scenarios as possible including in-distribution (ID) data and Out-Of-Distribution (OOD) data \citep{rajak_ex-nnqmd_2021}. Nonstoichiometric compounds material is useful in new material property research. They exhibit different properties such as conductivity, magnetism, catalytic nature, and other unique solid-state properties, which have important technological applications \citep{Rogacheva12, geonu_kim_benchmark_2023, orlov_nanoscale_2015, rogacheva_non-stoichiometry_2006, dubey_stoichiometric_2019, kostenko_vacancy_2021}. 
Therefore, it would be worthwhile if a model with high generalization can be learned and applied to different atom compositions. Meanwhile, it is necessary and crucial to improve the stability of GNNFF models, especially in the OOD dataset.
Another important challenge is how to evaluate the stability of a GNNFF model in MD simulation. Since the metrics in training process cannot be applied in MD simulation, the current measurement of GNNFF model is insufficient for stability evaluation \citep{geonu_kim_benchmark_2023}.

In this paper, we target on improving the stability of GNNFF models in MD simulation especially over OOD datasets, and propose a GNN feature correlation based method in GNNFF training. Our key contributions are as follows: 
\begin{itemize} [leftmargin=10pt] 
  \setlength{\itemsep}{0pt}
  \setlength{\parsep}{0pt}
  \setlength{\parskip}{0pt}
\item We analyze the stability performance of GNNFF models with different structures and \textbf{reveal the negative relationship of feature correlation and stability} of MD simulation with GNNFF models.
\item To improve the stability of GNNFF models, we \textbf{design a loss function to reduce feature correlation} that can be applied during GNNFF model training.
\item	To alleviate the accuracy drop involved by extra loss function, we \textbf{design a scheduler to dynamically adjust loss coefficient} during the training.
\item	To better evaluate the effectiveness of our method, we \textbf{design an empirical metric} based on multiple physical values extracted from simulation results.
\end{itemize}

\section{Related Work}

\subsection{GNNFF models}
\textbf{Graph Neural Network based Force Field}: Given an atomic system with $n$ atoms, each atom has an atomic number and position $ \bm r_i\in\mathbb{R}^{n \times 3}$, and Force Field (FF) models learn from the interactions of atoms to predict the system potential energy $E$ and force $ \bm {F_i}$ for each atom. Typically, the forces on each particle are obtained as $  \bm {F_i} = -\partial E / \partial  \bm {r}_i $ \citep{fu_forces_2022}. In GNNFFs, atoms are considered to be nodes and the interaction or bonds between two atoms are considered to be edges. An edge is built when the distance of two atoms is less than a predefined cutoff threshold. GNNFF learns knowledge from atoms' spatial information like distances, angles between atom pairs, and dihedral of atom groups. The accuracy of FF model is usually evaluated by Energy MAE (EMAE) and Force MAE (FMAE) per-atom, with the unit of meV/atom and meV/Å.

\textbf{NequIP} \citep{batzner_e3-equivariant_2022} is an E(3)-equivariant Message Passing Network employing E(3)-equivariant convolutions for interactions of geometric tensors. It achieves state-of-the-art accuracy on a challenging and diverse set of molecules and materials with remarkable data efficiency.
\textbf{Allegro} \citep{musaelian_learning_2023} is a local interaction based-FF model. It predicts the energy as a function of the final edge embedding rather than the node embeddings. All the pairwise energies are summed to obtain the total energy of the system. Allegro shows high accuracy and great scalability with its local interaction architecture.
\textbf{GemNet} \citep{gasteiger_gemnet_2021} is a Message Passing Network based on directed edge embeddings and two-hop message passing. GemNet and its variants shows high accuracy in OC20 \citep{ocp_dataset_2020} leaderboard but lower scalability than Allegro.

\begin{figure}[t]
  \centering
  \includegraphics[width=0.48\textwidth]{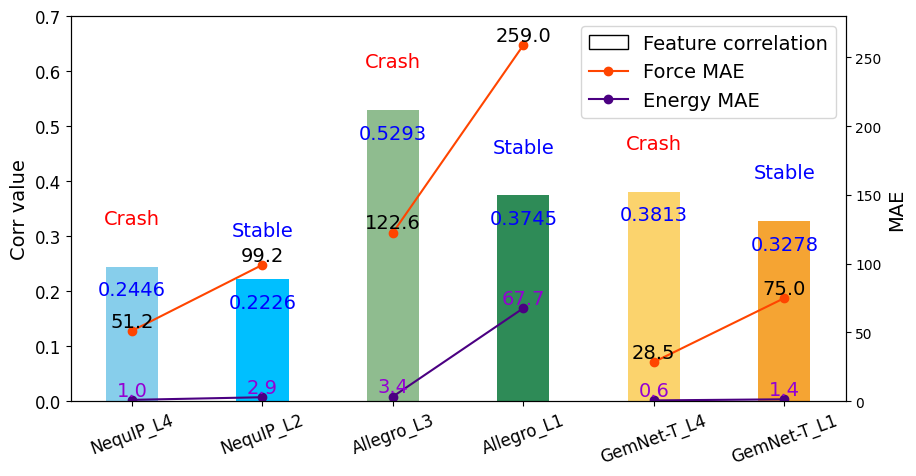}
  \caption{Feature correlation and MD simulation stability of GNNFF models of different layers}
  \label{fig_corr}
  \vspace*{-1.0\baselineskip}
\end{figure}

\subsection{MD simulation stability}

Recently, the stability of MD simulation when using MLFF/GNNFF models to describe atomic interaction is actively discussed in the field. MLFFs may produce unstable prediction result when the learned force field is not robust to the under-sampled data distribution \citep{orlov_nanoscale_2015, rogacheva_non-stoichiometry_2006, dubey_stoichiometric_2019, kostenko_vacancy_2021}. The simulation can enter nonphysical states that would never occur in a realistic simulation and eventually MD simulation will end up as system crash as shown in the left side of Figure \ref{fig_nonphysical}. 

Accordingly, some methods have been proposed to relieve the MD simulation instability issue in MLFF area in recent years. For example, active learning \citep{vandermause_--fly_2020, xie_bayesian_2021, vandermause_active_2022, xie_uncertainty-aware_2023} can be used to improve the accuracy and stability of the MLFF model by increasing the quality and diversity of the training dataset. When the uncertainty in model predictions exceeds a specified threshold, the model is retrained using newly generated training data. However, generating new training data needs additional DFT calculation, which is time and resource consuming. Therefore, even though many methods have emerged to accelerate the active learning process, retraining MLFF model with active learning is still costly and less scalable.

There are already some existing methods dealing with the generalization issue in neural networks, including dropout \citep{Dropout}, weight decay \citep{weightdecay}, early stopping \citep{earlystop}, flatter loss landscapes \citep{keskar_large-batch_2017, dziugaite_computing_2017, jiang_fantastic_2020, vita_data_2023}, etc. But only flatten loss landscapes are disccussed in improving stability of GNNFFs.
\citeauthor{vita_data_2023} used loss entropy to quantify the flatness of the loss landscape, and they used different training parameters to increase the loss entropy and thus improve the MD stability.
\citeauthor{foret_sharpness-aware_2021} approximated the minimization of sharpness by Sharpness-Aware Minimization (SAM), and successfully improved the out-of-sample error of the model on the MLFF model.
\citeauthor{ibayashi_allegro-legato_2023} improves MD stability of Allegro model by SAM in training process. The results show that it can expand the simulation time of Allegro model. However, these methods come at the cost of some training overhead and accuracy loss. For example, Allegro-Legato increases the training time of Allegro model by 75\%, and decreases FMAE from 10.7 mev/Å to 11.6 mev/Å.

Besides, since the traditional metric (FMAE/EMAE) cannot measure MLFFs' stability in real MD simulation scenarios, some other metrics are proposed to quantify MD stability: Time to Failure \citep{ibayashi_allegro-legato_2023}, Wright's Factor (WF), and Jensen-Shannon Divergence (JSD) \citep{rajak_ex-nnqmd_2021} of Radial Distribution Function (RDF) analysis. Time to Failure roughly measures stability with simulation time but misses other important physical metrics in MD simulation. WF and JSD need additional reference data generated from DFT which requires lots of computation resources in simulation.

\begin{figure}[t]
  \centering
  \includegraphics[width=0.45\textwidth]{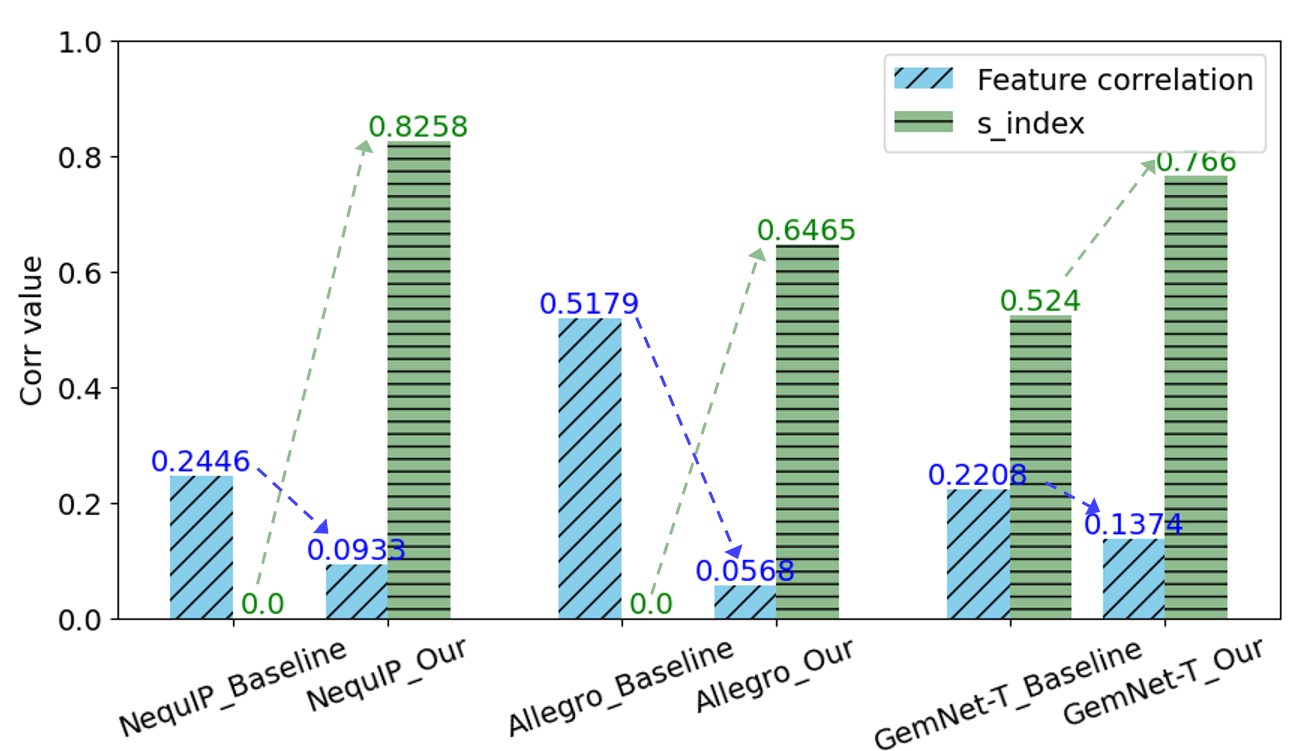}
  \caption{Feature correlation and MD simulation stability of GNNFF models with our method}
  \label{fig_corr_sindex}
  \vspace*{-1.0\baselineskip}
\end{figure}

\section{Methodology}
\label{headings}

Our method is inspired by the performance deterioration of deep GNNs \citep{li_deeper_2018}. The potential issue of deep GNNs lies in over-smoothing \citep{zhao_pairnorm_2020, chen_measuring_2020} and over-correlation \citep{jin_feature_2022}. Over-smoothing indicates the learned node representations become highly indistinguishable when stacking too many GNN layers. Over-correlation indicates that deeply stacking GNN layers renders the learned feature dimensions highly correlated. High correlation indicates high redundancy and less information encoded by the learned dimensions, which can harm the generalization and performance of downstream tasks.

Therefore, in order to understand the trends of stability performance with different GNN architectures over ID and OOD datasets, we have trained NequIP, Allegro and GemNet-T models with different layers over the hafnium oxide (HfO) ID dataset released in \citep{geonu_kim_benchmark_2023}. HfO is typically used as a high-k material and a crucial ferroelectric material in complementary metal-oxide-semiconductor technology, showing great potential for emerging electronics applications. The ID training dataset comprises 96 atoms, exhibiting a 1:2 ratio of 32 Hf atoms to 64 O atoms, and OOD datasets exhibiting a 1:1.1 and 1:1.55 ratio of Hf atoms to O atoms. OOD dataset is added in our benchmark because different types of atom compositions for HfO material exists in real application scenario.

We benchmarked the simulation stability with the trained model over both ID and OOD datasets for 40,000 steps in LAMMPS \footnote {It is equivalent to 10ps of MD simulation with 0.25fs simulation step in LAMMPS}. The result in Table \ref{tab_diff_layers} shows shallow GNNFFs with one/two layers provide more stable MD simulation trajectories than deep GNNFFs with three/four layers even the latter show better accuracy in ID dataset. "Fail" in Table \ref{tab_diff_layers} means MD simulation completed, but results are not physical; "Crash@s384" means MD simulation crashed at step 384. We noticed that larger models with more layers fail to simulate stably while smaller models with less layers succeed because large models are not fully trained with the limited dataset. However, small models suffers lower MAE.

\begin{table*}
  \centering
  \small
  \begin{tabular}{ccccccc}
    \toprule
    \multicolumn{1}{c}{\multirow{2}{*}{Model}} & \multicolumn{1}{c}{\multirow{2}{*}{Layers}} & \multicolumn{1}{c}{\multirow{2}{*}{\makecell{FMAE\\ (meV/Å)}}} & \multicolumn{1}{c}{\multirow{2}{*}{\makecell{EMAE\\ (meV/atom)}}} & \multicolumn{3}{c}{Simulation stability with atom compositions}\\
    \cline{5-7}
                            &  &  &  & Hf:O = 1:1.1 & Hf:O = 1:1.55 & Hf:O = 1:2.0 \\
    \midrule
    \multirow{2}{*}{NequIP}   & 2  & 99.2  & 2.9  &	Stable     & Stable       & Stable \\
                              & 4  & 51.2  & 1.0  & Crash@s384 & Crash@s6698  & Stable \\
    \midrule
    \multirow{2}{*}{Allegro}  & 1  & 259.0 & 67.7 &	Stable     & Stable       & Stable \\
                              & 3  & 122.6 & 3.4  & Crash@s150 & Crash@s301   &	Stable \\
    \midrule
    \multirow{2}{*}{GemNet-T} & 1  & 75.0  & 1.4  & Stable     & Stable       & Stable \\
                              & 4  & 28.5  & 0.6  & Fail       & Fail         & Stable \\
    \bottomrule
  \end{tabular}
  \caption{MD Simulation stability test result of GNNFFs with different number of layers}
  \label{tab_diff_layers}
\end{table*}

To better understand the stability results in Table \ref{tab_diff_layers}, we measured the edge feature correlation of each model. Figure \ref{fig_corr} shows models with lower feature correlation value are more stable during MD simulation than those with higher feature correlation value. After analyzing the feature correlation of these GNNFF models, we found the over-correlated features reduce the generalization of deep GNNs. And thus cause instability of deep GNNFF models over OOD datasets. Therefore, reducing feature correlation of deep GNNFF models in training can improve the generalization and increase the stability of MD simulation.

\subsection{Overview}
Our method aims to reduce feature correlation of models and thereby improve GNNFF model generalization. We add an extra loss function in GNNFF model training to punish high feature correlation, and no modification on model architecture is involved. Therefore, this method can be applied to any GNNFF models.
Figure \ref{fig_overview} shows the whole workflow of our method, which contains three main components: (1) feature correlation based loss function, (2) dynamic loss coefficient scheduler, and (3) stability index evaluator. Correlation loss function focuses on reducing feature correlation in the back propagation process. Loss coefficient scheduler dynamically changes the loss coefficient of correlation loss and avoids model from only focusing on reducing feature correlation and ignoring optimizing accuracy. The stability index evaluator will evaluate the stability of model from multiple aspects during MD simulation.

Our method contains three steps: Step 1 and 2 is applied for GNNFF model training; Step 3 is applied for MD simulation. The workflow is described as follows:
\begin{itemize} [leftmargin=10pt] 
  \setlength{\itemsep}{0pt}
  \setlength{\parsep}{0pt}
  \setlength{\parskip}{0pt}

\item Step 1. Output edge features of each GNN layer and compute correlation loss with edge features.

\item Step 2. Compute correlation loss coefficient with dynamic scheduler and apply it to correlation loss.

\item Step 3. Output MD simulation snapshots with intervals and evaluate simulation stability with evaluator.

\end{itemize}

In \cite{jin_feature_2022} the metric to measure over-correlation and the feature correlation based loss function to alleviate over-correlation was proposed. However, unlike \cite{jin_feature_2022}, we use edge features instead of node features to better relieve over-correlation issue in GNNFFs. In GNNFF models, edge features are propagated and aggregated layer by layer and finally accumulated to get atom and system potential energy, so all edge features are the smallest components in the potential energy, which is critical to ensure the stability of MD simulation. Based on this, we choose to reduce the correlation degree between edge features. Besides, in Ablation Study section we discussed the effectiveness of using edge features instead of node features.

\subsection{Feature correlation calculation} \label{sec_corr}
Supposing that a GNN model has $L$ layers and each layer will produce a set of edge features to pass the message to the next layer, we denote the edge features as $X_1,\ldots ,X_l,\ldots,X_L$. Each $X_l$ has shape $[f,dim]$, where $f$ is the number of edges and $dim$ is the dimension of edge features. We define feature correlation as the correlation value between each dimension of edge feature, so the correlation matrix $Corr_l \in \mathbb{R} ^{dim \times dim}$ is shaped like $[dim, dim]$, and can be calculated from the $l$-th GNN layer.
$Corr_l [k,j]$ is the element located at row $k$ and column $j$ of $Corr_l$, which means the correlation value between feature dimension $k$ and feature dimension $j$, and can be calculated by:

\begin{equation}
  \setlength {\abovedisplayskip}{1pt} %
  Corr_l[k,j]=|\rho(X_l (:,k), X_l (:,j))|,
  \label{eq1}
\end{equation}
where $\rho (X, Y)$ is the Pearson correlation coefficient \citep{benesty_pearson_2009}, which measures linear correlation between column vectors $X$ and $Y$. In ideal case, we expect that the correlation coefficient between any pair of different feature dimensions to be 0, which implies there is no linear dependency between them.

For equivariant GNNFFs \citep{batzner_e3-equivariant_2022}, features are geometric objects that comprise a direct sum of irreducible representations of the O(3) symmetry group. Therefore, we need to do extra processing on features to select $1o$ features to calculate the feature correlation.

Computing all edge features of all atoms from all training samples is time-consuming, so we randomly sample $\sqrt{f}$ edges from all $f$ edges in a sample to calculate correlation value. The number of multiplications to compute the covariance matrix of edge features decreases from $dim^2 f$ to $dim^2 \sqrt{f}$.

\subsection{Correlation loss function}

We expect the feature correlation of each layer can be as low as possible, so the target is to optimize $Corr_l$ to an identity matrix $Corr_{target}$. The loss function is: 
\begin{equation}
  loss_{corr}^l = \frac{\sum|Corr_l-Corr_{target}|} {dim(dim-1)}.
  \label{eq4}
\end{equation}
Finally, we sum $loss_{corr}^l$ of all layers to $loss_{corr}$, and our final optimizing target is to minimize $loss_{corr}$:
\begin{equation}
  loss_{corr}=\sum {loss_{corr}^l}
  \label{eq5}
\end{equation}
To measure the correlation value of a model on a specified dataset, only the correlation matrix $Corr_L$ at the last layer of the model is taken. If we suppose there are $B$ samples in the dataset, the final correlation value is:
\begin{equation}
  Corr= \sum_{b = 1}^{B}Corr_L^b.
  \label{eq3}
\end{equation}

\subsection{Dynamic coefficient scheduler}
Combining the two loss functions (force and energy) is tricky since focusing on one metric may lead to performance degradation on the other, not to mention the extra correlation loss involved by our method. Thus, it is necessary to balance stability, energy accuracy and force accuracy. Therefore, we propose a dynamic coefficient scheduler to balance those objectives:

\vspace*{-1.0\baselineskip}
\begin{equation}
  c_{corr}^t = c_{max}-\frac{c_{max}-c_{min}}{2} \cdot ({1 + cos (\frac{t}{t_{cycle}}\cdot\pi)})
  \label{eq6}
\end{equation}

Before each training epoch starts, the current loss coefficient $c_{corr}^t$ is updated. $[c_{min},c_{max}]$ is the range of correlation loss coefficient during training; $t_{cycle}$ is the update epoch cycle interval, and $t$ is the current epoch; $c_{min}$ $c_{max}$, and $t_{cycle}$ are hyperparameters to be set before training. In our experiments, $c_{min}=0$, $c_{max}=0.1,t_{cycle}=100$.

Our correlation loss coefficient scheduler is similar to cyclic cosine annealing learning rate scheduler \citep{loshchilov_sgdr_2017}, but our coefficient scheduler is gradually increasing instead of decreasing in one cycle due to the priority given to force and energy accuracy. In the early stage of training process, the model can quickly converge to the minimum. If the correlation loss coefficient is high, the correlation loss acting as a regular term will put the model into a poor local minimum. Similarly, periodically restarting the coefficient can help jump out of current local minimum and find a lower local minimum to improve accuracy.
The total loss is

\vspace*{-1.0\baselineskip}
\begin{equation}
  loss=c_f \cdot loss_f+c_e \cdot loss_e+c_{corr}^t \cdot loss_{corr}
  \label{eq7}
\end{equation}

Finally, backward calculation is proceeded and gradients are updated according to the loss value calculated by Eq.\ref{eq7} until the model is fully trained.
Empirically, $c_{max}$ should not be set bigger than $c_f$ and $c_e$ to avoid accuracy drop. For example, if $c_f$ and $c_e$ are 1 respectively, 1 or 0.1 is preferred for $c_{max}$.

\subsection{Empirical metric to evaluate MD stability}
 Physical values and atom information in simulation results (such as temperature, force, number of atoms, length of bonds, etc.) can be used to evaluate the stability of a model in simulation experiments. However, evaluating with multiple non-consecutive values would be confused for users. Therefore, we propose a unified metric to quantify the stability performance with all the meaningful simulation values.

The empirical metric considers atom number, forces' abnormality and distance between pairs of atoms to quantify the stability of GNNFF models over a dataset in MD simulation. Moreover, system temperature is proportional to kinetic energy, which explains why unstable simulation always shows unusually high temperatures.

Simulations usually crash because model predicts forces that are extremely huge, and so atoms fly out of simulation space and lost. Then crash happens because of unmatched atom number. Our metric, the stability index, takes all the above situations into account, as shown in Eq.\ref{eq11}.

\begin{equation}
s_{index}=\frac{1}{num} \sum_{n=1}^{num}S_{index}^n
\label{eq11}
\end{equation}

Specifically, to estimate the typical and physically correct values of MD simulation, first we run MD simulation for a certain number of steps with the trained model in LAMMPS framework. Then, we dump the simulation trajectory data with a fixed simulation step interval such as saving for every 100 steps. The saved trajectory data should include atom number $N$, atom positions $ \bm r$ and temperatures $T$. After simulation, $num$ snapshots are saved. For each snapshot, a stability index $s_{index}$ should be calculated and accumulated together. 
Second, we calculated $r_{min}$, the minimum distance between atoms for different atomic species pairs using atom positions $ \bm r$ which is just dumped. Just like RDF value, with total C atomic species in a system, combining pairwise species can get $ \frac {1}{2} C(C+1)$ total number of the atom pair composition, so we need to calculate $ \frac {1}{2} C(C+1)$ sets of $r_{min}$ values.
To calculate the stability index of the $n$-th snapshot, take current atom number $N_n$  and initial atom number $N_0$, set simulation temperature $T_{set}$, current temperature $T_n$, current minimum distance $r_{{min}_n}$ and last minimum distance $r_{{min}_{(n-1)}}$ into Eq.\ref{eq12}

\vspace*{-1.0\baselineskip}
\begin{equation}
S_{index}^n=\left({\frac{N_n}{N_0}}\right)^\alpha \cdot \left({\frac{T_{set}}{T_n}}\right) ^\beta \cdot \prod \limits_{i=1}^{\frac{1}{2}C(C+1)}\left({r_{min_n}^i -r_{min_{n-1}}^i}\right),
\label{eq12}
\end{equation}
where $\alpha$ is the scale factor for atom number, and $\beta$ is the scale factor for temperature. In our experiments we have used $\alpha$ = 1, and $\beta$ = 1/4. The higher $s_{index}$ is, the more stable the simulation is. 

\section{Experimental validation}

\label{others}

We conducted a number of experiments to show the effectiveness of our method, evaluated model accuracy over ID and OOD datasets and stability performance in LAMMPS simulation, and estimated the overhead of our method. All the LAMMPS simulations are conducted with Langevin thermostat and with timestep equals 2.5fs. Since "fix langevin" command does not perform time integration (it only modifies forces to effect thermostatting) \cite{thompson_lammps_2022}, we use a separate time integration with microcanonical NVE ensemble to actually update the velocities and positions of atoms using the modified forces. We also conducted ablation study to show the effectiveness of each component in our method. All the experiments were done on the Supercomputing Center, where each server node has 8 NVIDIA A100-SXM4-80GB GPU connected in series via NVLink. The software versions are: PyTorch 12.1 and CUDA 11.4.

\subsection{Evaluation on GNNFF models}

 We take NequIP, Allegro and GemNet-T models to evaluate the performance with correlation method applied. We use the default model configurations from original proposed paper. We fit our GNNFF models with the newly released HfO dataset designed for semiconductor advanced material called the SAMD23 dataset \citep{geonu_kim_benchmark_2023}.
To assess the accuracy of GNNFF models, we used EMAE and FMAE over ID test datasets, similarly to \citep{geonu_kim_benchmark_2023}. Furthermore, in order to better evaluate the stability, we use the proposed metric $s_{index}$ to quantify the stability of MD simulations (see Table \ref{tab_performance}).

\subsection{Accuracy}

FMAE and EMAE columns in Table \ref{tab_performance} shows the accuracy of the GNNFF models on HfO ID dataset of baseline and our method. Corr value is the correlation value of last layer calculated by equation \ref{eq1}. GemNet-T model with our method achieves lower Force MAE. However, NequIP and Allegro model trained with our method suffer accuracy loss. This is because GemNet-T used more geometric features and interaction information applied with full graph structure while NequIP only used a pair of atom interaction information and Allegro only used local geometric information. Therefore, our method has less impact on the accuracy of GemNet-T model than that of NequIP and Allegro.

\subsection{Stability}

The stability experiments are conducted over ID and OOD datasets with different Hf:O ratios. We expect to perform stable MD simulation over both ID and OOD with one unified model rather than train different models for different compositions. We perform 40,000 steps of simulation with temperature of 1,200K, 1,800K and 2,400K in LAMMPS with baseline model and the model trained by our method respectively.

\begin{figure}
  \centering
  \begin{minipage}{1\linewidth}
  \centering

  \subfigure[Baseline step 20]{
    \includegraphics[width=0.3\linewidth,height=0.3\linewidth]{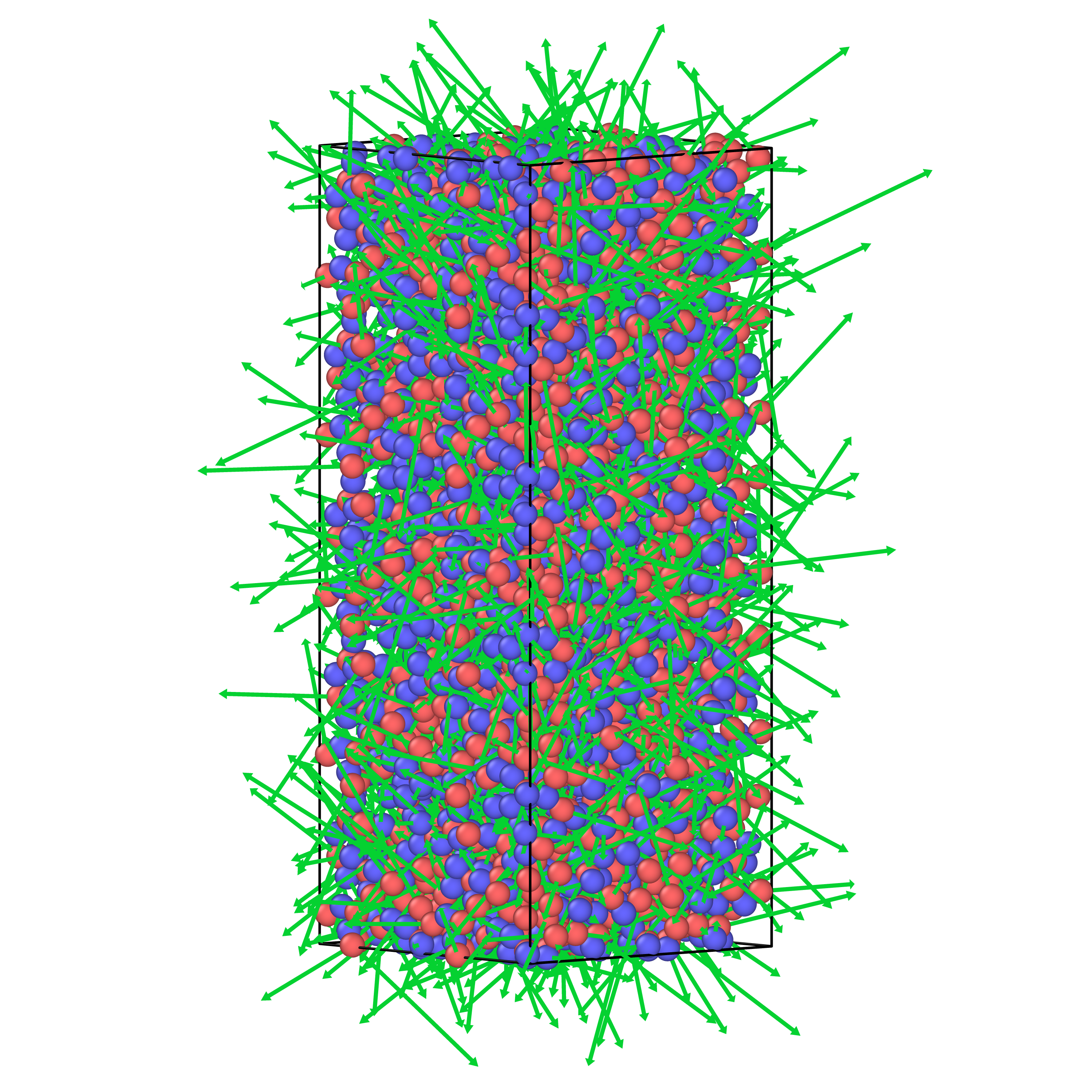}
  }
  \subfigure[Baseline step 70]{
    \includegraphics[width=0.3\linewidth,height=0.3\linewidth]{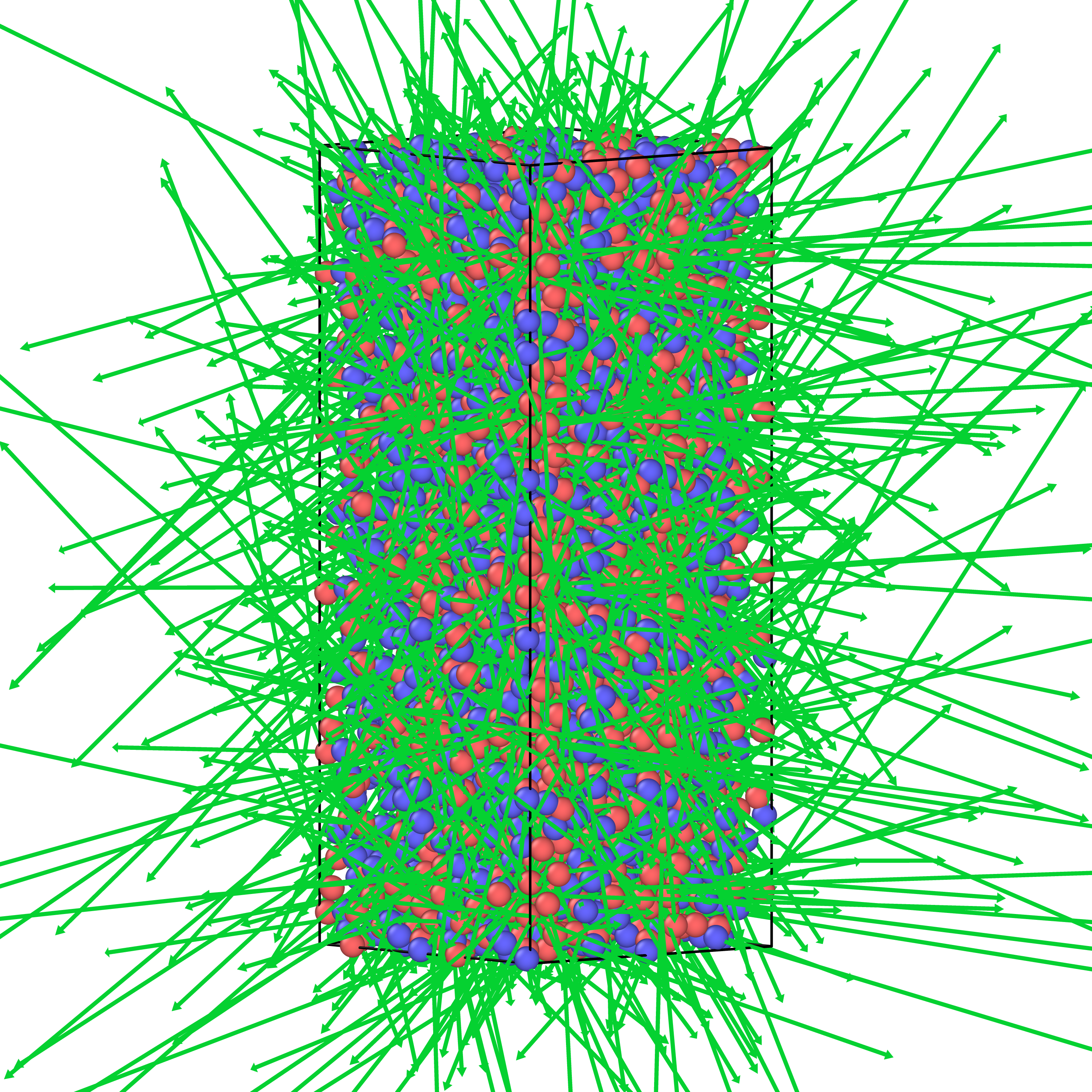}
  }
  \subfigure[Baseline step 150]{
    \includegraphics[width=0.3\linewidth,height=0.3\linewidth]{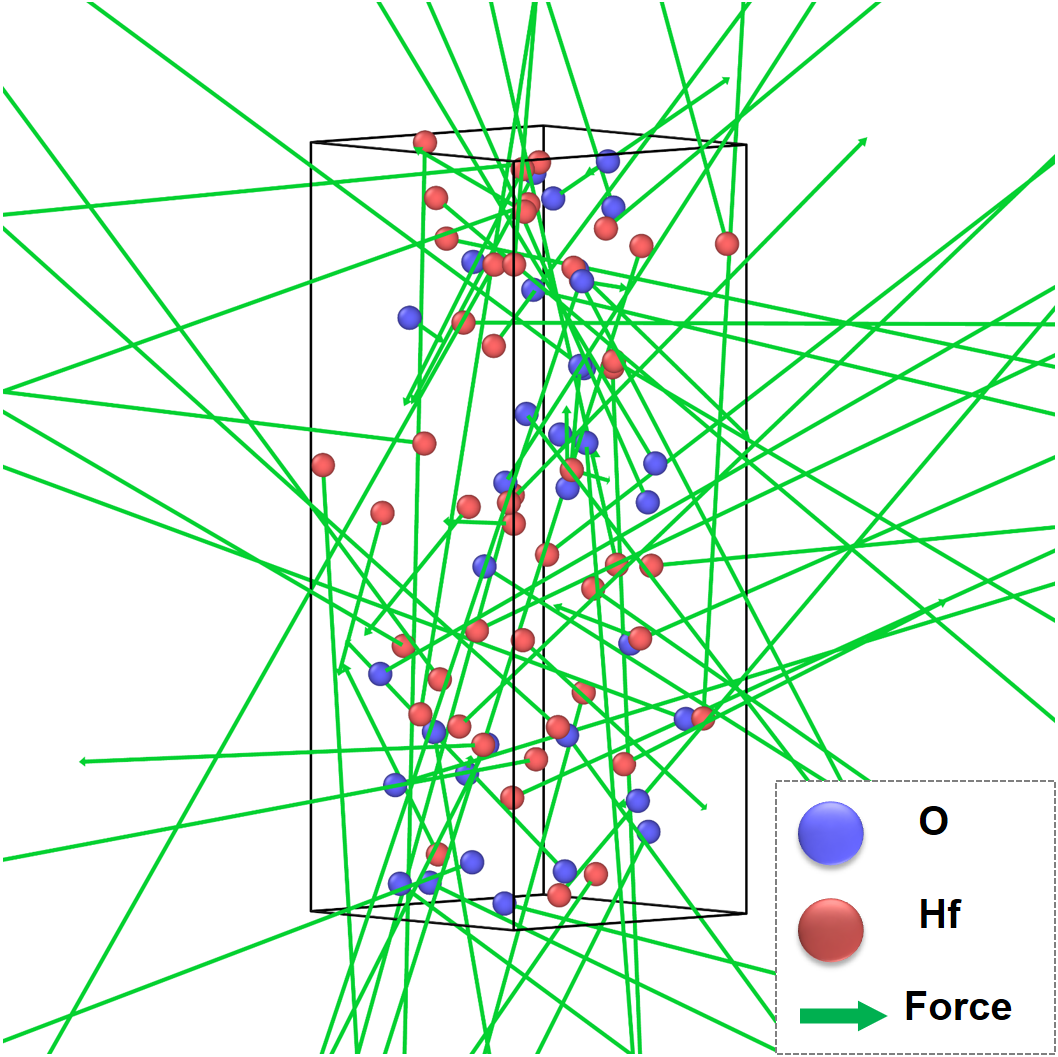}
  }
  \end{minipage}
  \begin{minipage}{1\linewidth}
  \centering
  \vspace*{-0.5\baselineskip}
  \subfigure[Ours step 20]{
    \includegraphics[width=0.3\linewidth,height=0.3\linewidth]{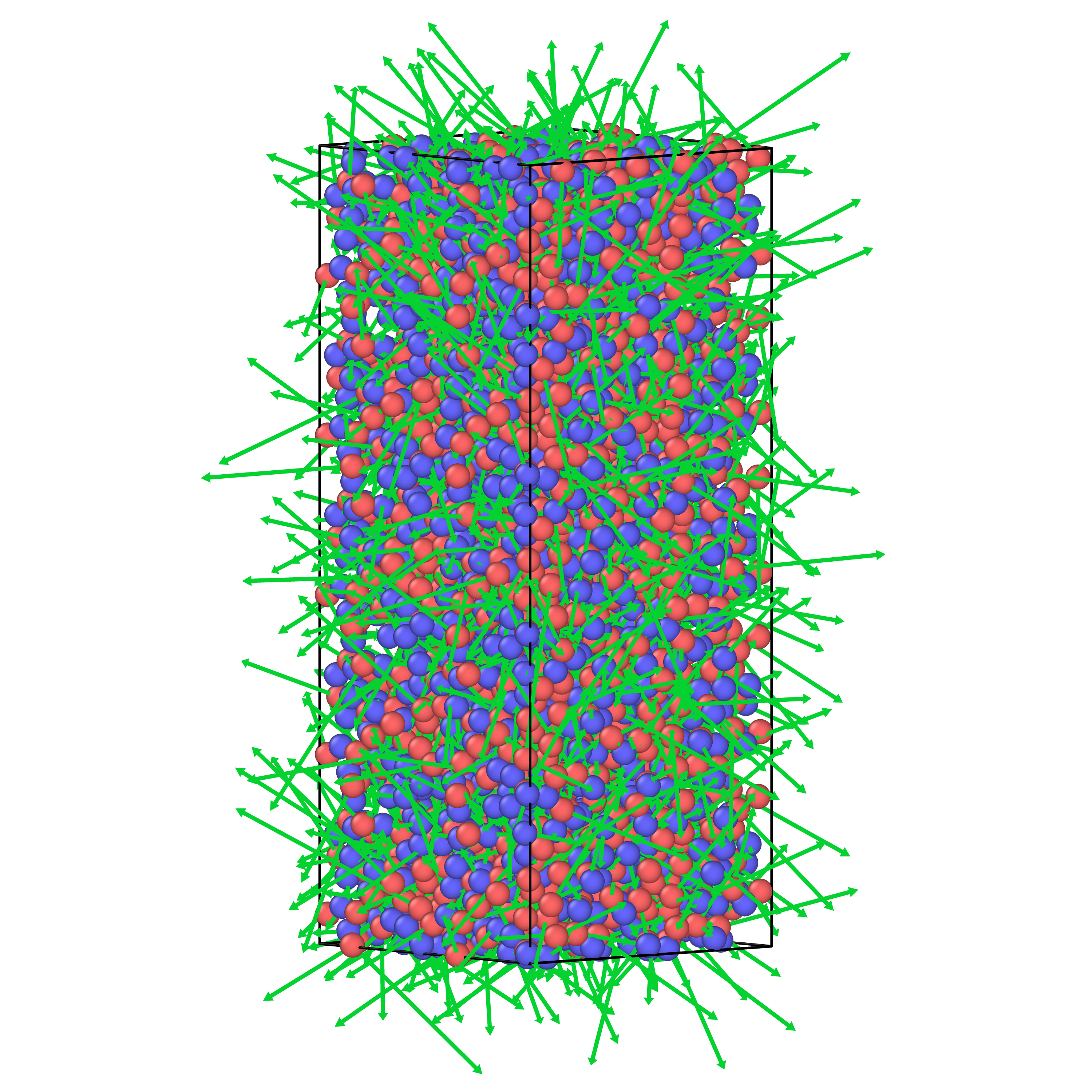}
  }
  \subfigure[Ours step 70]{
    \includegraphics[width=0.3\linewidth,height=0.3\linewidth]{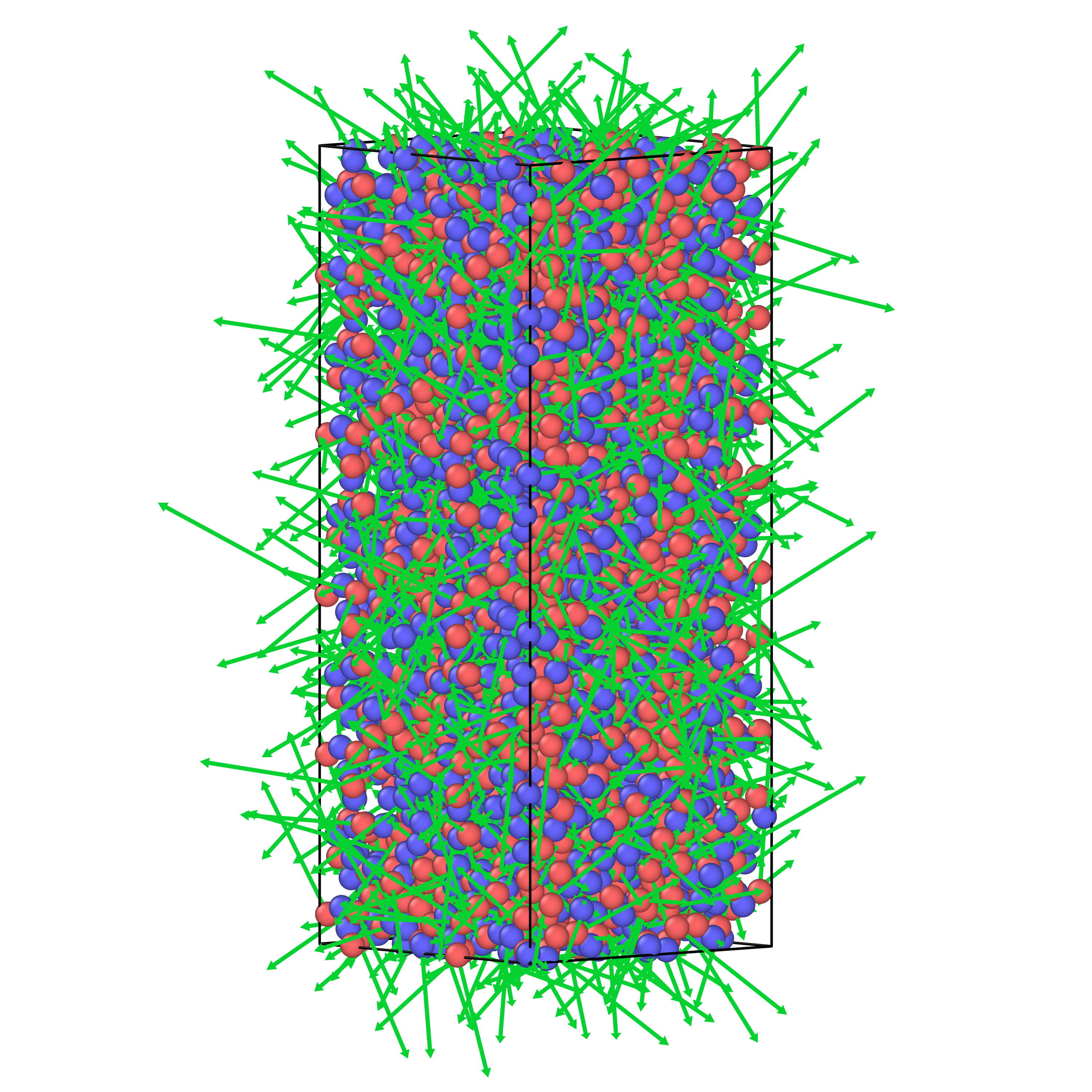}
  }
  \subfigure[Ours step 150]{
    \includegraphics[width=0.3\linewidth,height=0.3\linewidth]{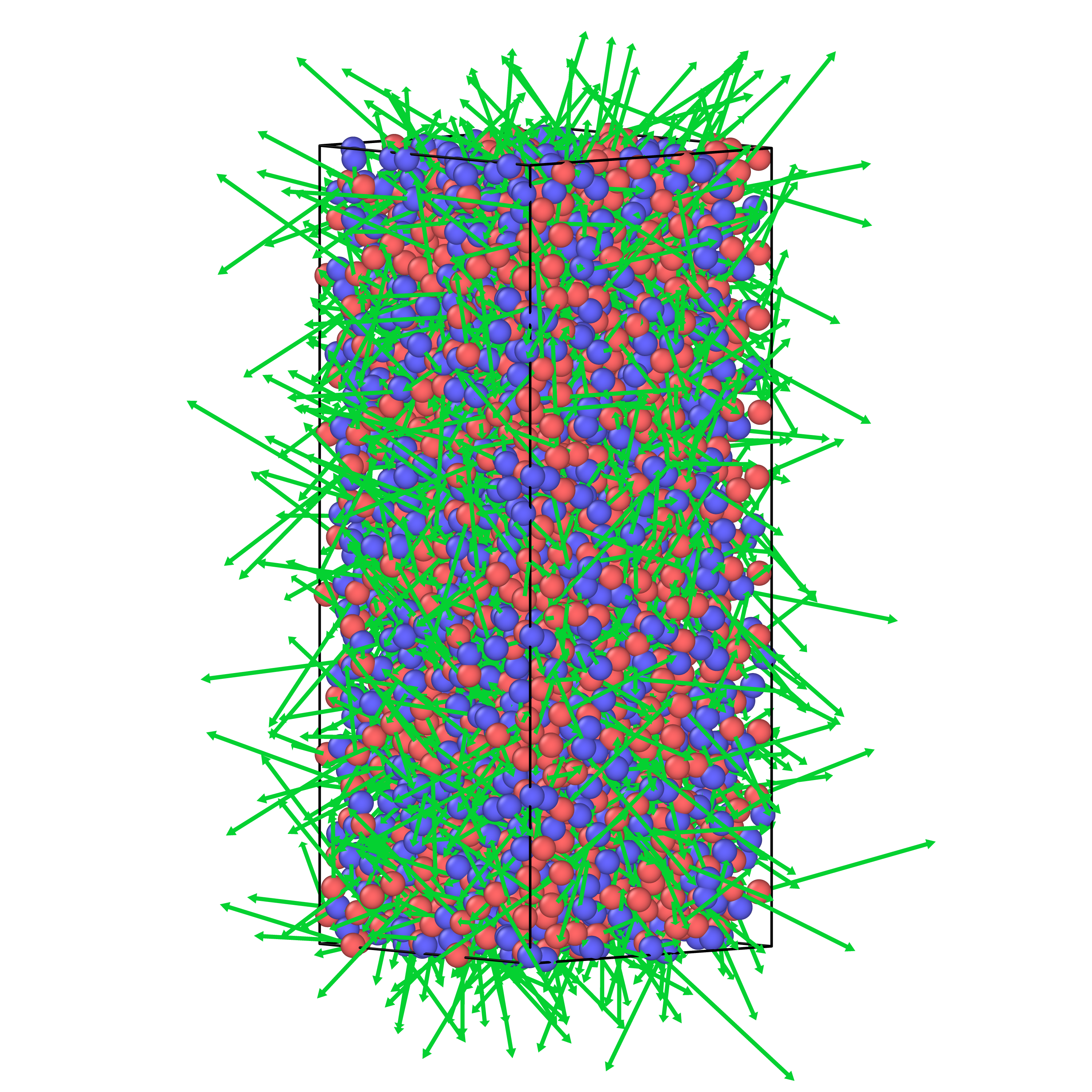}
  }
  \vspace*{-1.0\baselineskip}

\end{minipage}
  \caption{Simulation result of Hf:O =1:1.1 with baseline Allegro model and optimized Allegro model. (a)-(c) shows the simulation status of step 20, 70 and 150 with baseline Allegro model. (d)-(f) shows the simulation status of step 20, 70 and 150 with optimized Allegro model.}

  \label{fig_allegro_atom}

\end{figure}

\begin{figure*}[t]
  \vspace*{-0.5\baselineskip}
  \centering

  \subfigure[Baseline]{
    \includegraphics[width=0.43\linewidth,height=0.27\linewidth]{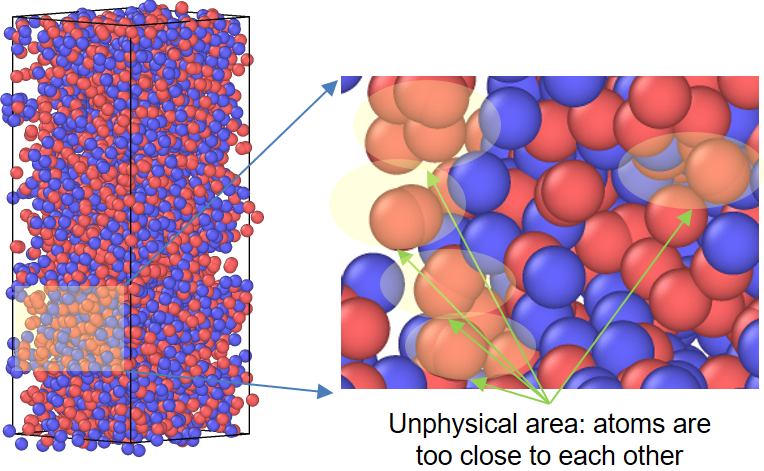}
  }
  \subfigure[Optimized]{
    \includegraphics[width=0.43\linewidth,height=0.27\linewidth]{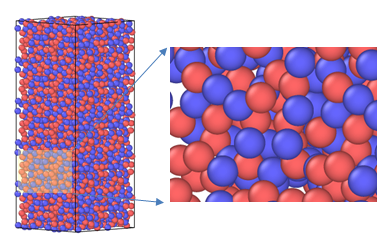}
  }
  \vspace*{-1.0\baselineskip}
  \caption{Simulation result of Hf:O =1:1.1 with baseline and optimized GemNet-T model}
  \label{fig_gemnet_atom}
  \vspace*{-1.0\baselineskip}
\end{figure*}

\begin{figure*}
  \centering
  \includegraphics[width=0.8\textwidth]{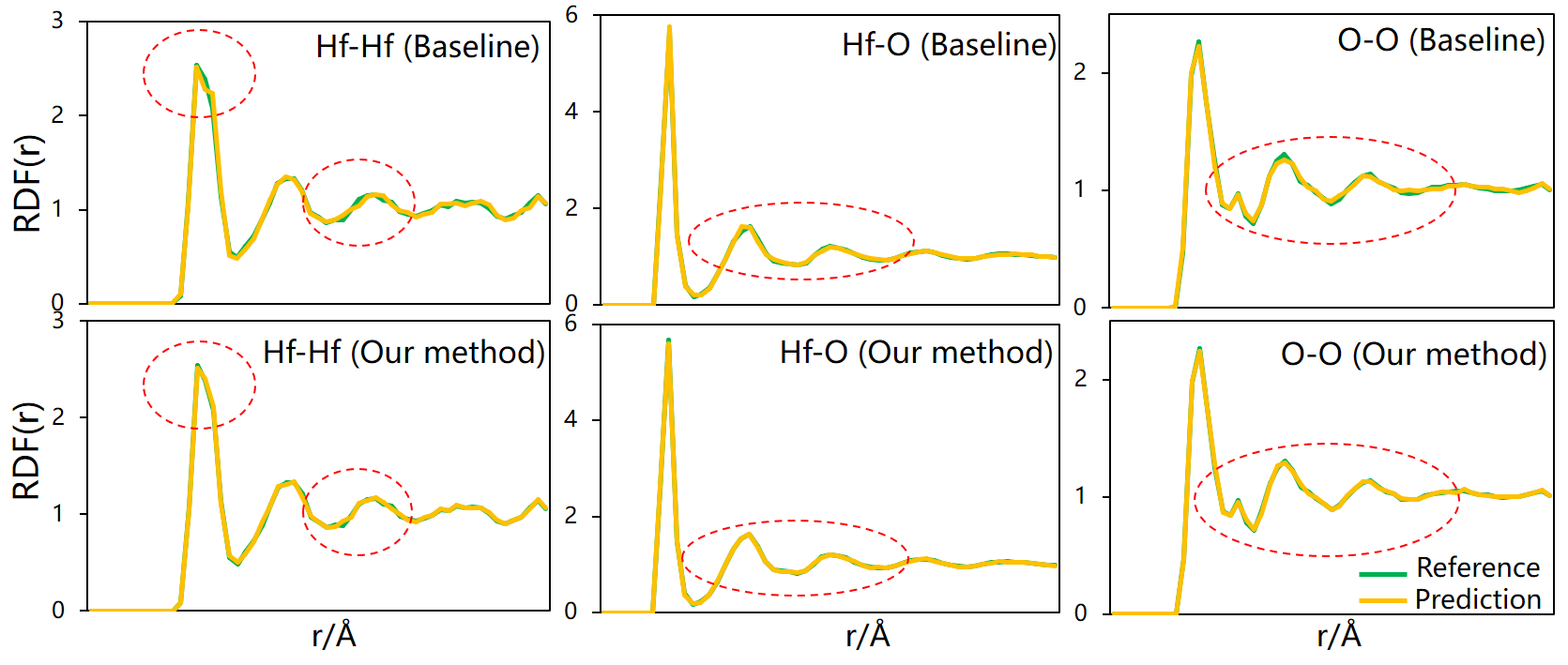}
  \caption{RDFs of GemNet-T model over HfO dataset}
  \label{fig_rdf_gemnet}
  \vspace*{-1\baselineskip}
 \end{figure*}

As shown in Table \ref{tab_performance}, baseline NequIP model can only successfully perform MD simulation with ID dataset, while other two cases with OOD dataset get $s_{index}=0$, which indicating system crash because of lost atoms. NequIP model trained with our method can complete MD simulation with both ID and OOD dataset. Furthermore, all cases achieve reasonable physical values during simulation. We also get similar result for Allegro model. With our method, the simulation time can be extended from 0.03ps to 10ps as shown in Figure \ref{fig_nonphysical}. As shown in Figure \ref{fig_allegro_atom}, optimized Allegro model gets more reasonable and stable force values in all simulation steps while baseline model crashed because of unreasonable force values (red spheres represent Hf atom, purple spheres represent O atom and green-colored vectors represent the force of the atoms during MD simulation). For GemNet-T model, the baseline completes the simulation, but the distance between close atoms shows abnormality as shown in Figure \ref{fig_gemnet_atom}. GemNet-T model trained with our method can run simulation successfully with all 3 cases and achieve reasonable physical values including forces and atom distances for all the cases. We list the result from the simulation temperature of 1,200K, but we see the similar trend with the simulation temperature of 1,800K and 2,400K. Figure \ref{fig_rdf_gemnet} shows the RDF curves for HfO (1:2) dataset of GemNet-T model baseline and optimized with our method. We can see that RDF curve with optimized model involves less noisy compared with the baseline. We have got similar RDF results for NequIP and Allegro models.

All above shows that our proposed method remove all non-physical results from the simulated structures, and thus provide stable MD simulation results. 
More information on stability experiments are shown in section \ref{hfo_evaluation} and detailed physical metric values of MD simulation are presented in Table \ref{tab_stability}.

\begin{table*}[t]
  \small
  \centering
  \begin{tabular}{ccccccc}
    \toprule
    \multicolumn{1}{c}{\multirow{2}{*}{Model}} & \multicolumn{1}{c}{\multirow{2}{*}{Corr value}} & \multicolumn{1}{c}{\multirow{2}{*}{\makecell{FMAE\\ (meV/Å)}}} & \multicolumn{1}{c}{\multirow{2}{*}{\makecell{EMAE \\ (meV/atom)}}} & \multicolumn{3}{c}{$s_{index}$ over different Hf:O}\\
    \cline{5-7}
     &  &  &  & 1:1.1 & 1:1.55 & 1:2.0 \\
    \midrule
    NequIP (Baseline)  & 0.2446  & 51.2  & 1.0  &	0   & 0     & 0.8490 \\
    NequIP (Our)   & 0.0933  & 61.6 (+10.4)  & 1.3 (+0.3)  & 0.8258 & 0.8264  & 0.8490 \\
    \midrule
    Allegro (Baseline)  & 0.5179  & 122.6 & 3.4 &	0   & 0     & 0.8494 \\
    Allegro (Our)  & 0.0568  & 134.0 (+11.4) & 4.2 (+0.8)  & 0.6465 & 0.8138   &	0.8484 \\
    \midrule
    GemNet-T (Baseline) & 0.2208  & 20.5  & 0.3  & 0.524   & 0.5824     & 0.8496 \\
    GemNet-T (Our) & 0.1374  & 19.7 (-0.8)  & 0.4 (+0.1)  & 0.7660  & 0.7845  & 0.8496 \\
    \bottomrule
  \end{tabular}
  \caption{MD Simulation stability test result of different GNNFFs}
  \label{tab_performance}
\end{table*}

\begin{table}[H]
  \small
  \centering
  \begin{tabular}{cccc}
    \toprule
    Model        & NequIP            & Allegro          &GemNet-T    \\
    \midrule
    Baseline     & 1702.7            & 365.5            & 1,880.3      \\
    Optimized    & 1719.2            & 377.5            & 1,899.7      \\
    Overhead     &  +1\%             & +3\%             & +1\%         \\
    \bottomrule
  \end{tabular}
  \vspace*{-0.5\baselineskip}
  \caption{Comparison of computational overhead incurred by our method in seconds/epoch.}
  \label{tab_overhead}
\end{table}

\begin{table}
  \centering
  \small
  \vspace*{-0.5\baselineskip}
  \begin{tabular}{lllcccc}
    \toprule
    Method       & FMAE      &EMAE      & Crash  & $s_{index}$  \\
    \midrule
    Baseline   &  51.2 & 1.0             & Y      & 0  \\
    \midrule
    Edge ($0e$) & 75.1 & 1.9      & Y       & 0\\
    \midrule
    Edge ($1o + 0e$)  & 76.6 & 1.9     & N        & 0.7189\\
    \midrule
    Edge ($1o$)   &  61.6 & 1.3            & N       & 0.8258 \\
    \midrule
    Node  &  52.6 & 1.1      & Y      & 0\\
    \bottomrule
  \end{tabular}
  \caption{MD simulation result of different NequIP models by using Node and Edge features to calculate Correlation Over Hf:O=1:1.1 }
  \label{tab_nequip_feature}
\end{table}

\subsection{Ablation study} \label{ablation_study}

\textbf{Computation overhead.}
To reduce the computation overhead involved in correlation calculation, only a small portion of sampled features are used to compute correlation, so there is little overhead even an extra loss function is involved in training. In our experiments we randomly choose features of 1,024 edges among the total edges to calculate the correlation. Results in Table \ref{tab_overhead} show there is only up to 3\% extra overhead in NequIP, Allegro and GemNet-T.

\textbf{Correlation calculation.}
Different from Allegro and GemNet-T model, features in NequIP are geometric objects that comprise a direct sum of irreducible representations of the O(3) symmetry group \citep{batzner_e3-equivariant_2022}. Therefore, we tried the following four types of combination with different orders and parities of features: a) mixing up all edge feature with different parities and rotation orders; b) only taking $0e$ ($l$=0 and parity is even) features; c) only taking $1o$ ($l$=1 and parity is odd) features; d) summing the correlations of $0e$ and $1o$ features. Besides, we reduce correlation of node features to see if stability is improved. Table \ref{tab_nequip_feature} shows reducing correlation of $1o$ features can achieve higher accuracy and better stability in MD simulation. The result comparison of reducing correlation of edge features and node features also shows reducing correlation of edge features are more useful which proves that the information in edge features are more crucial for GNNFFs.

\textbf{Coefficient scheduler.}
 We experimented with two types of scheduler: linear scheduler and cosine scheduler. The former uses a linear increasing coefficient with training epochs; the latter uses a cycling increasing coefficient with a fixed epoch cycle.
The result in Table \ref{tab_scheduler} shows reducing feature correlation with both linear and cosine scheduler can help Allegro model to improve the generalization and achieve more stable MD simulation over OOD dataset. Furthermore, we can see cosine scheduler can achieve lower energy MAE than linear scheduler.

\section{Summary and Limitations}
This paper presents a method to improve GNNFF model's generalization and stability in MD simulation. Our method reduces GNN feature correlation by adding a correlation loss and dynamically scheduled coefficient. Evaluation results verify that our method can improve the simulation stability for GNNFF models both on ID and OOD datasets with less than 3\% computational overhead. Besides, this paper proposes a new metric to reveal the robustness of MD simulation with more physical information in simulation trajectory data. 

\textbf{Limitations.} Admittedly, the main limitation of the present study is that the motivation and studies are base on the GNN structures for MD tasks only, therefore the effectiveness of our method is validated over GNNFFs. Also, main focus of our work is semiconductor applications, where long-term simulations are critically needed. In the future, we aim to validate the generalization of our approach on numerous other GNNFF datasets and assessing the impact of model and data scaling. 

\clearpage
\begin{quote}
  \begin{small}
    \bibliography{stability_refs}
  \end{small}
\end{quote}
\clearpage

\clearpage
\appendix
\onecolumn
\section{Appendix / supplemental material}
\newcounter{sfigure}
\newcounter{stable}
\setcounter{sfigure}{1}
\setcounter{stable}{1}
\renewcommand{\thefigure}{A\arabic{sfigure}}
\renewcommand{\thetable}{A\arabic{stable}}

\subsection{Correlation calculation} \label{sec_features}
\textbf{Features used to calculate correlation.}
 There are two important features in Allegro model: edge features and environment features. We train Allegro by reducing correlation of only edge features and both edge features with environment features respectively. Both two cases using the same correlation coefficient equals 0.1 and fixed correlation coefficient. Result in Table \ref{tab_allegro_feature} shows reducing the correlation of both edge features and environment features at the same time can achieve better stability in MD simulation. And also using both two features achieves lower FMAE than only using edge features. Therefore, we can say that reducing more the correlation of more features is more helpful to improve the generalization of GNNFF models. 

 For NequIP models, in which geometric objects that comprise a direct sum of irreducible representations of the O(3) symmetry group \citep{batzner_e3-equivariant_2022}. Therefore, we do extra process with features in NequIP to get the feature correlation. We tried the following four types of combination with different orders and parities: a) mixing up all feature with different parities and rotation orders; b) only taking $0e$ ($l$=0 and parity is even) features; c) only taking $1o$ ($l$=1 and parity is odd) features; d) summing the correlations of $0e$ and $1o$ features. The result in Table \ref{tab_nequip_feature_2} shows reducing correlation of $1o$ features can achieve higher accuracy and better stability in MD simulation.

\begin{figure}[hb]
  \centering

    \centering
    \includegraphics[width=0.48\textwidth]{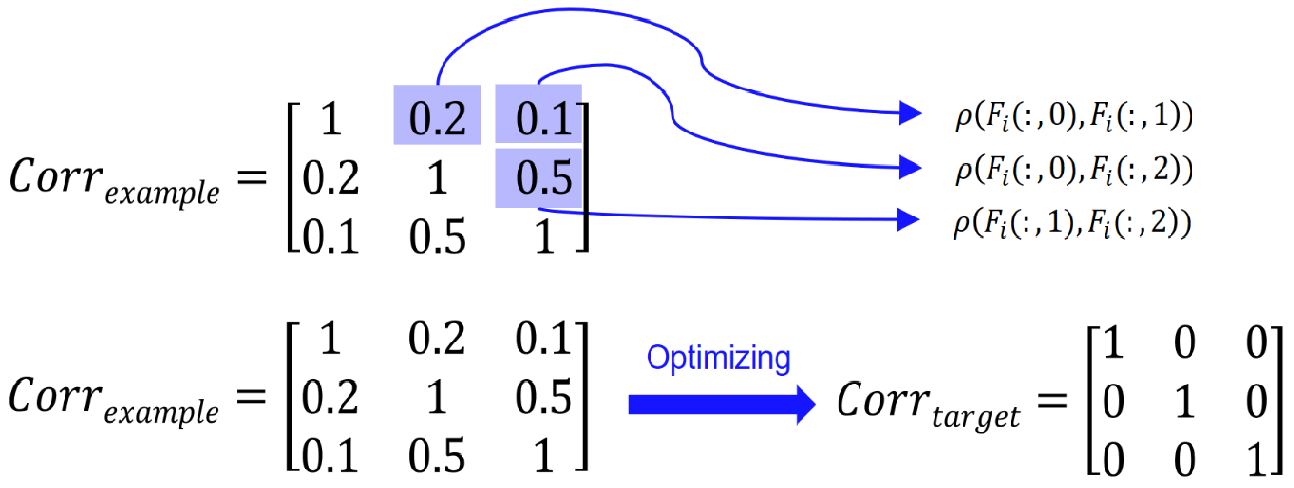}
    \caption{An example of correlation matrix and correlation target}
    \label{fig_corr_matrix}
    \addtocounter{sfigure}{1}
\end{figure}

 Figure \ref{fig_corr_matrix} shows an example of correlation matrix and the optimization target.
 
 Table \ref{tab_scheduler} shows the details of MD simulation result of different Allegro models by using different coefficient schedulers in our method.

\begin{figure*}[hb]
  \centering
  \begin{minipage}{0.9\linewidth}

  \subfigure[Baseline step 20]{
    \includegraphics[width=0.23\textwidth,height=0.23\textwidth]{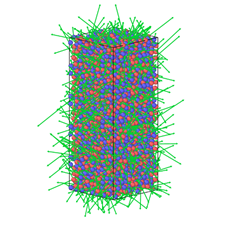}
  }
  \subfigure[Baseline step 390]{
    \includegraphics[width=0.23\linewidth,height=0.23\linewidth]{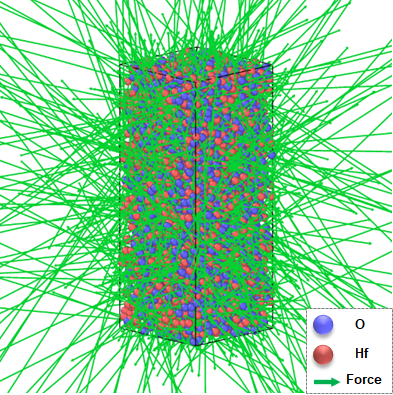}
  }
  \subfigure[Optimized step 20]{
    \includegraphics[width=0.23\linewidth,height=0.23\linewidth]{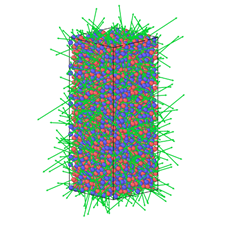}
  }
  \subfigure[Optimized step 390]{
    \includegraphics[width=0.23\linewidth,height=0.23\linewidth]{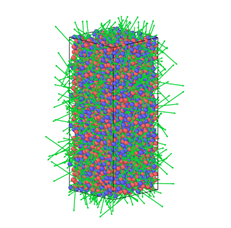}
  }
\end{minipage}
\caption{Simulation status of Hf:O =1:1.1 with baseline and optimized NequIP Model. (a)~(b) shows the simulation status of step 20 and 390 with baseline NequIP model. The simulation crashed at step 395 because of force abnormal. (c)~(d) shows the simulation status of step 20 and 390 with optimized NequIP model}
\label{fig_nequip_atom}
\addtocounter{sfigure}{1}

\end{figure*}

\begin{table}
  \centering
  \small
  \begin{tabular}{lllllll}
    \toprule
    Method       & Metric          & Hf:O       &System Crash  & Atom lost & Temp. (K) & $s_{index}$   \\
    \midrule
    \multirow{3}{*}{Baseline}     & \multirow{3}{*}{\makecell{FMAE: 122.6 \\EMAE: 3.4}}          & 1:1.10      & Y    & Y   &2e+27  & 0       \\
                                  &            & 1:1.55      & Y    & Y   &7e+29  & 0       \\
                                  &            & 1:2.00      & N    & Y   &1,259  & 0.8494  \\
    \midrule
    \multirow{3}{*}{Corr (Edge)}  & \multirow{3}{*}{\makecell{FMAE: 140.9 \\ EMAE: 4.1}}    & 1:1.10   & Y   & Y   &7e+11  & 0       \\
                                  &           & 1:1.55      & Y    & Y   &4e+11  & 0        \\
                                  &           & 1:2.00      & N    & Y   &1,248  & 0.8637   \\
    \midrule
    \multirow{3}{*}{Corr (Edge + Env.)}  & \multirow{3}{*}{\makecell{FMAE: 132.0\\ EMAE: 4.9}}  & 1:1.10      & N    & Y   &2e+61  & 0\\
                                         &   & 1:1.55      & N    & Y   &1,396  & 0.7929\\
                                         &   & 1:2.00      & N    & Y   &1,250  & 0.8634\\
    \bottomrule
  \end{tabular}
  \caption{MD stability test by reducing correlation of different features on Allegro}
  \label{tab_allegro_feature}
\end{table}

\begin{table}
  \centering
  \small
  \begin{tabular}{lllcccc}
    \toprule
    Method       & Metric           & Hf:O       &System  Crash  & Atom lost & Temp. (K) & $s_{index}$  \\
    \midrule
    \multirow{3}{*}{Baseline}     & \multirow{3}{*}{\makecell{FMAE: 51.2 \\ EMAE: 1.0}}         & 1:1.10      & Y    & N   &2e+8  & 0  \\
                                  &            & 1:1.55      & Y    & N   &2e+8  & 0  \\
                                  &            & 1:2.00      & N    & N   &1,262  & 0.8490   \\
    \midrule
    \multirow{3}{*}{Edge ($0e$)}  & \multirow{3}{*}{\makecell{FMAE: 75.1 \\ EMAE: 1.9}}  & 1:1.10     & Y    & N   &5e+9  & 0\\
                         &   & 1:1.55     & N    & N   &1,359  & 0.7929\\
                         &   & 1:2.00     & N    & N   &1,264  & 0.8634\\ 
    \midrule
    \multirow{3}{*}{Edge ($1o + 0e$)}  & \multirow{3}{*}{\makecell{FMAE: 76.6 \\EMAE: 1.9}} & 1:1.10      & N    & N   &1,507  & 0.7189\\
                              &    & 1:1.55     & N    & N   &1,321  & 0.8311\\
                              &    & 1:2.00     & N    & N   &1,261  & 0.8327\\
    \midrule
    \multirow{3}{*}{Edge ($1o$)}   & \multirow{3}{*}{\makecell{FMAE: 61.6 \\ EMAE: 1.3 }}        & 1:1.10     & N     & N   &1,270  & 0.8258 \\
                          &          & 1:1.55     & N     & N   &1,279  & 0.8264 \\
                          &          & 1:2.00     & N     & N   &1,262  & 0.8490   \\
    \midrule
    \multirow{3}{*}{Node}  & \multirow{3}{*}{\makecell{FMAE: 52.6 \\ EMAE: 1.1}}  & 1:1.10     & Y    & N   &2e+9  & 0\\
                         &   & 1:1.55     & Y    & N   &2e+8  & 0\\
                         &   & 1:2.00     & N    & N   &1,262  & 0.8490\\
    \bottomrule
  \end{tabular}
  \addtocounter{stable}{1}
  \caption{MD simulation result of different NequIP models by using Node and Edge features to calculate Correlation }
  \label{tab_nequip_feature_2}
\end{table}

\begin{table}
  \centering
  \small
  \begin{tabular}{lllcccc}
    \toprule
    Method       & Metric           & Hf:O       &System Crash  & Atom lost & Temp. (K) &$s_{index}$  \\
    \midrule
    \multirow{3}{*}{Baseline}  & \multirow{3}{*}{\makecell{FMAE: 122.6 \\EMAE: 3.4}}  & 1:1.10   & Y  & Y  &2e+27  & 0 \\
                                  &      & 1:1.55      & Y    & Y   &7e+29  & 0   \\
                                  &      & 1:2.00      & N    & N   &1,259  & 0.8494     \\
    \midrule
    \multirow{3}{*}{Fixed}    & \multirow{3}{*}{\makecell{FMAE: 132.0 \\ EMAE: 4.9}}    & 1:1.10     & N     & Y   &2e+61  & 0 \\
                              &     & 1:1.55     & N     & N   &1,396  & 0.7929     \\
                              &     & 1:2.00     & N     & N   &1,250  & 0.8634   \\
    \midrule
    \multirow{3}{*}{Linear}  & \multirow{3}{*}{\makecell{FMAE: 133.0\\ EMAE: 4.4}} & 1:1.10     & N    & N   &1,785  & 0.7091    \\
                             &    & 1:1.55     & N    & N   &1,346  & 0.8272\\
                             &    & 1:2.00     & N    & N   &1,249  & 0.8635\\
    \midrule
    \multirow{3}{*}{Cosine}  & \multirow{3}{*}{\makecell{FMAE: 134.0 \\EMAE: 4.2}} & 1:1.1      & N    & N   &1,795  & 0.6465   \\
                             &    & 1:1.55     & N    & N   &1,347  & 0.8138\\
                             &    & 1:2.00     & N    & N   &1,266  & 0.8484\\
    \bottomrule
  \end{tabular}
  \addtocounter{stable}{1}
  \caption{MD simulation result of different Allegro models by using different coefficient schedulers in our method.}
  \label{tab_scheduler}
\end{table}

\subsection{Evaluation on HfO dataset} \label{hfo_evaluation}
We list the more detail physical values (atom numbers, simulation temperature, force values, atom distances) in MD simulation with baseline models and our models as shown in Table \ref{tab_stability}. Result shows that models optimized with our method can show more reasonable physical values in MD simulation with OOD HfO atoms. 

Figure \ref{fig_nequip_atom} shows the MD simulation status when using baseline and optimized NequIP model. Red sphere represents Hf atom, purple sphere represents O atom and green-colored vectors represent the force of the atoms during MD simulation. We can see that optimized NequIP model get more stable and reasonable force predictions compared with baseline NequIP model.

\begin{table*}
  \resizebox*{\textwidth}{!}{
  \begin{tabular}
    {ccccccccccc}
    \toprule
    Method & \makecell{Corr\\value}  & Hf:O    &\makecell{System \\Crash}  & \makecell{Atom \\lost} & Temp. (K) & \makecell{Force \\Abn.} & \makecell{Min dis \\(Hf-Hf)} & \makecell{Min dis\\(Hf-O)}& \makecell{Min dis\\ (O-O)} & $s_{index}$ \\
    \midrule
    \multirow{3}{*}{\makecell{NequIP \\(Baseline)}}     & \multirow{3}{*}{0.2446}     & 1:1.10      & Y    & N   &2e+8   & 202.725  & 2.5 & 0.5 & 1.8 & 0       \\
         &        & 1:1.55      & Y    & N   &2e+8   & 1e+6     & 0.2 & 0.3 & 0.2 & 0       \\
         &        & 1:2.00      & N    & N   &1,262  & 55.071   & 2.7 & 1.7 & 2.1 & 0.8490  \\
    \midrule
    \multirow{3}{*}{\makecell{NequIP \\ (Opt.)} }   & \multirow{3}{*}{0.0933}    & 1:1.10     & N     & N   &1,270  & 51.450    & 2.3 & 1.7 & 2.1 & 0.8258 \\
         &        & 1:1.55     & N     & N   &1,279  & 51.783    & 2.3 & 1.7 & 2.1 & 0.8264 \\
         &        & 1:2.00     & N     & N   &1,262  & 55.107    & 2.7 & 1.7 & 2.1 & 0.8490 \\
    \midrule
    \multirow{3}{*}{\makecell{Allegro \\ (Opt.)}}  & \multirow{3}{*}{0.5179}   & 1:1.10     & Y    & Y   &2e+27  & 4e+15    & 1.9 & 1.2 & 2.1 & 0       \\
        &        & 1:1.55     & Y    & Y   &7e+29  & 5e+15    & 2.3 & 1.6 & 1.8 & 0       \\
        &        & 1:2.00     & N    & N   &1,259  & 51.504   & 2.7 & 1.7 & 2.1 & 0.8494  \\
    \midrule
    \multirow{3}{*}{ \makecell{Allegro \\ (Opt.)}}  & \multirow{3}{*}{0.0568}   & 1:1.1     & N    & N   &1,795  & 55.887  & 2.0 & 1.7 & 1.3 & 0.6465 \\
         &       & 1:1.55    & N    & N   &1,347  & 53.116  & 2.4 & 1.7 & 2.0 & 0.8138 \\
         &       & 1:2.00    & N    & N   &1,266  & 51.355  & 2.7 & 1.7 & 2.1 & 0.8484 \\
    \midrule
    \multirow{3}{*}{\makecell{GemNet-T \\ (Baseline)}}  & \multirow{3}{*}{0.2208}   & 1:1.10     & N    & N   &1,282  & 51.708  & 0.8 & 1.6 & 2.1 & 0.524  \\
         &      & 1:1.55     & N    & N   &1,265  & 51.884  & 1.0 & 1.7 & 2.0 & 0.5824 \\
         &      & 1:2.00     & N    & N   &1,258  & 55.013  & 2.7 & 1.7 & 2.1 & 0.8496 \\
    \midrule
    \multirow{3}{*}{\makecell{GemNet-T \\(Opt.)}}  & \multirow{3}{*}{0.1374}  & 1:1.1     & N    & N   &1,270  & 52.012  & 2.0 & 1.7 & 2.0 & 0.7660 \\
         &      & 1:1.55    & N    & N   &1,260  & 52.527  & 2.1 & 1.7 & 2.1 & 0.7845 \\
         &      & 1:2.00    & N    & N   &1,258  & 55.032  & 2.7 & 1.7 & 2.1 & 0.8496 \\
    \bottomrule
  \end{tabular}}  
  \addtocounter{stable}{1}
  \caption{MD simulation result of baseline GNNFF models and optimized model with our method. (Temp=1,200K)}
  \label{tab_stability}
\end{table*}

\subsection{GNNFF Model Configurations} \label{model_config}
Table \ref{tab_NequIP_config} lists the hyper-parameters used for NequIP model training. Table \ref{tab_Allegro_config} lists the hyper-parameters used for Allegro model training.  Table \ref{tab_GemNet-T_config} lists the hyper-parameters used for GemNet-T model training.

\begin{table*}

  \centering
  \begin{tabular}{ll}
  \toprule
  NequIP hyperparameters  & Value  \\ 
  \midrule
  BesselBasis\_trainable  & true    \\
  PolynomialCutoff\_p    & 6       \\
  avg\_num\_neighbors    & auto    \\
  r\_max                 & 6.0     \\
  l\_max                 & 2       \\
  parity                 & true    \\
  num\_layers            & 4       \\
  invariant\_layers      & 2       \\
  invariant\_neurons     & 64      \\
  nonlinearity\_type     & gate    \\
  resnet                 & false   \\
  nonlinearity\_gates    & e: silu \\
                          & o: tanh \\
  nonlinearity\_scalars  & e: silu \\
                          & o: tanh \\
  num\_basis              & 8       \\
  num\_features           & 32      \\
  use\_sc                 & true    \\ 
  \bottomrule
  \end{tabular}
  \addtocounter{stable}{1}
  \caption{NequIP model architecture configuration}
  \label{tab_NequIP_config}
  \end{table*}
  
  \begin{table*}
  \centering
  \begin{tabular}{ll}
  \toprule
  Allegro hyperparameters                    & Value                \\ \midrule
  BesselBasis\_trainable                     & true                 \\
  PolynomialCutoff\_p                        & 6                    \\
  avg\_num\_neighbors                        & auto                 \\
  r\_max                                     & 6.0                  \\
  l\_max                                     & 2                    \\
  parity                                     & o3\_restricted       \\
  num\_layers                                & 3                    \\
  env\_embed\_multiplicity                   & 16                   \\
  embed\_initial\_edge                       & true                 \\
  two\_body\_latent\_mlp\_latent\_dimensions & {[}32, 32, 32, 32{]} \\
  two\_body\_latent\_mlp\_nonlinearity       & silu                 \\
  two\_body\_latent\_mlp\_initialization     & uniform              \\
  latent\_mlp\_latent\_dimensions            & {[}32{]}             \\
  latent\_mlp\_initialization                & uniform              \\
  latent\_resnet                             & true                 \\
  env\_embed\_mlp\_nonlinearity              & null                 \\
  env\_embed\_mlp\_initialization            & uniform              \\
  edge\_eng\_mlp\_latent\_dimensions         & {[}32{]}             \\
  edge\_eng\_mlp\_nonlinearity               & null                 \\
  env\_embed\_mlp\_initialization            & uniform              \\
  edge\_eng\_mlp\_latent\_dimensions         & {[}32{]}             \\
  edge\_eng\_mlp\_nonlinearity               & null                 \\
  edge\_eng\_mlp\_initialization             & uniform              \\ \bottomrule
  \end{tabular}
  \addtocounter{stable}{1}
  \caption{Allegro model architecture configuration}
  \label{tab_Allegro_config}
  \end{table*}
  
  \begin{table*}
  \centering
  \begin{tabular}{ll}
  \toprule
  GemNet-T hyperparameters & Value                      \\ 
  \midrule
  activation               & silu                       \\
  cbf                      & spherical\_harmonics       \\
  cutoff                   & 6.0                        \\
  direct\_forces           & false                      \\
  emb\_size\_atom          & 128                        \\
  emb\_size\_bil\_trip     & 64                         \\
  emb\_size\_cbf           & 16                         \\
  emb\_size\_edge          & 128                        \\
  emb\_size\_rbf           & 16                         \\
  emb\_size\_trip          & 64                         \\
  envelope                 & exponent: 5                \\
                           & name: polynomial           \\
  extensive                & true                       \\
  max\_neighbors           & 50                         \\
  num\_after\_skip         & 1                          \\
  num\_atom                & 2                          \\
  num\_before\_skip        & 1                          \\
  num\_blocks              & 4                          \\
  num\_concat              & 1                          \\
  num\_radial              & 6                          \\
  num\_spherical           & 7                          \\
  otf\_graph               & true                       \\
  output\_init             & HeOrthogonal               \\
  rbf                      & spherical\_bessel          \\
  regress\_forces          & true                       \\
  use\_pbc                 & true                       \\ 
  \bottomrule
  \end{tabular}
  \addtocounter{stable}{1}
  \caption{GemNet-T model architecture configuration}
  \label{tab_GemNet-T_config}
\end{table*}
\end{document}